\documentclass[10pt,twocolumn,letterpaper]{article}

\usepackage{cvpr}
\usepackage{times}
\usepackage{epsfig}
\usepackage{graphicx}
\usepackage{amsmath}
\usepackage{amssymb}
\usepackage{makecell}
\usepackage{pdfpages}

\usepackage{multirow}
\usepackage{mathrsfs}
\usepackage{enumitem}
\usepackage{makecell}
\usepackage{tabulary}
\usepackage{pifont}
\usepackage{subfigure}

\usepackage{caption}

\usepackage[colorlinks=true,
    linkcolor=blue,pageanchor]{hyperref}

\cvprfinalcopy 

\begin{document}
\title{Pre-training for Video Captioning Challenge 2020 Summary\\\url{http://www.auto-video-captions.top/2020/}}

\author{Yingwei Pan, Jun Xu, Yehao Li, Ting Yao, and Tao Mei \\
{\normalsize\centering JD AI Research, Beijing, China}\\
{\tt\small \{panyw.ustc, junx1992, yehaoli.sysu, tingyao.ustc\}@gmail.com}
}

\twocolumn[{%
\renewcommand\twocolumn[1][]{#1}%
\maketitle
\begin{center}
\setlength\tabcolsep{3.6 pt}
   \centering
\captionof{table}{The leaderboard of top-3 submissions.}
\begin{tabular}{c|c|c|cccc}
\Xhline{1.0pt}
Rank  & Team Name  & Affiliation                                                                                                                                    & BLEU@4 & METEOR & CIDEr-D & SPICE \\ \hline
1 & \hyperlink{page.2}{Old Boys}{} & \begin{tabular}[c]{@{}c@{}}Tsinghua University,\\ Beijing University of Posts and Telecommunications,\\ Shanghai Ocean University\end{tabular} & 21.14  & 17.38  & 24.42   & 5.65  \\ \hline
2 & \hyperlink{page.3}{sysu-cs}{}    & Sun Yat-sen University                                                                                                                         & 20.41  & 17.02  & 23.80   & 5.39  \\ \hline
3 & \hyperlink{page.4}{IVIPC-King}{} & University of Electronic Science and Technology of China                                                                                       & 18.24  & 16.46  & 21.36   & 5.25  \\ \Xhline{1.0pt}
\end{tabular}
\label{leaderboard}
\end{center}
}]

\section{Challenge Introduction}
The Pre-training for Video Captioning Challenge is a Multimedia Grand Challenge in conjunction with ACM Multimedia 2020. The goal of this challenge is to offer a fertile ground for designing vision-language pre-training techniques that facilitate the vision-language downstream tasks (e.g., video captioning \cite{chen2019temporal,li2018jointly,pan2016jointly,pan2017video,Venugopalan:ICCV15,Yao:ICCV15} this year). Meanwhile, to further motivate and challenge the multimedia community, we provide a large-scale video-language pre-training dataset \cite{autogif2020} (namely ``\href{http://www.auto-video-captions.top/2020/dataset}{Auto-captions on GIF}'') for contestants to solve this challenging but emerging task.

Particularly, the contestants are asked to develop video captioning system based on Auto-captions on GIF dataset (as pre-training data) and the public MSR-VTT benchmark \cite{msrvtt} (as training data for downstream task). For the evaluation purpose, a contesting system is asked to produce at least one sentence of the test videos. The accuracy will be evaluated against human pre-generated sentence(s).

\section{Challenge Results}
Table \ref{leaderboard} details the results of top-3 submissions. We also attach to this document a copy of the technical reports submitted to the challenge.

{
	\bibliographystyle{ieee_fullname}
	\bibliography{pre-training}

\begin{thebibliography}{1}\itemsep=-1pt

\bibitem{chen2019temporal}
Jingwen Chen, Yingwei Pan, Yehao Li, Ting Yao, Hongyang Chao, and Tao Mei.
\newblock Temporal deformable convolutional encoder-decoder networks for video
  captioning.
\newblock In {\em AAAI}, 2019.

\bibitem{li2018jointly}
Yehao Li, Ting Yao, Yingwei Pan, Hongyang Chao, and Tao Mei.
\newblock Jointly localizing and describing events for dense video captioning.
\newblock In {\em CVPR}, 2018.

\bibitem{autogif2020}
Yingwei Pan, Yehao Li, Jianjie Luo, Jun Xu, Ting Yao, and Tao Mei.
\newblock Auto-captions on gif: A large-scale video-sentence dataset for
  vision-language pre-training.
\newblock {\em arXiv preprint arXiv:2007.02375}, 2020.

\bibitem{pan2016jointly}
Yingwei Pan, Tao Mei, Ting Yao, Houqiang Li, and Yong Rui.
\newblock Jointly modeling embedding and translation to bridge video and
  language.
\newblock In {\em CVPR}, 2016.

\bibitem{pan2017video}
Yingwei Pan, Ting Yao, Houqiang Li, and Tao Mei.
\newblock Video captioning with transferred semantic attributes.
\newblock In {\em CVPR}, 2017.

\bibitem{Venugopalan:ICCV15}
Subhashini Venugopalan, Marcus Rohrbach, Jeffrey Donahue, Raymond Mooney,
  Trevor Darrell, and Kate Saenko.
\newblock Sequence to sequence - video to text.
\newblock In {\em ICCV}, 2015.

\bibitem{msrvtt}
Jun Xu, Tao Mei, Ting Yao, and Yong Rui.
\newblock Msr-vtt: A large video description dataset for bridging video and
  language.
\newblock In {\em CVPR}, 2016.

\bibitem{Yao:ICCV15}
Li Yao, Atousa Torabi, Kyunghyun Cho, Nicolas Ballas, Christopher Pal, Hugo
  Larochelle, and Aaron Courville.
\newblock Describing videos by exploiting temporal structure.
\newblock In {\em ICCV}, 2015.

\end{thebibliography}
}


\begin{center}
\includegraphics[width=1\textwidth]{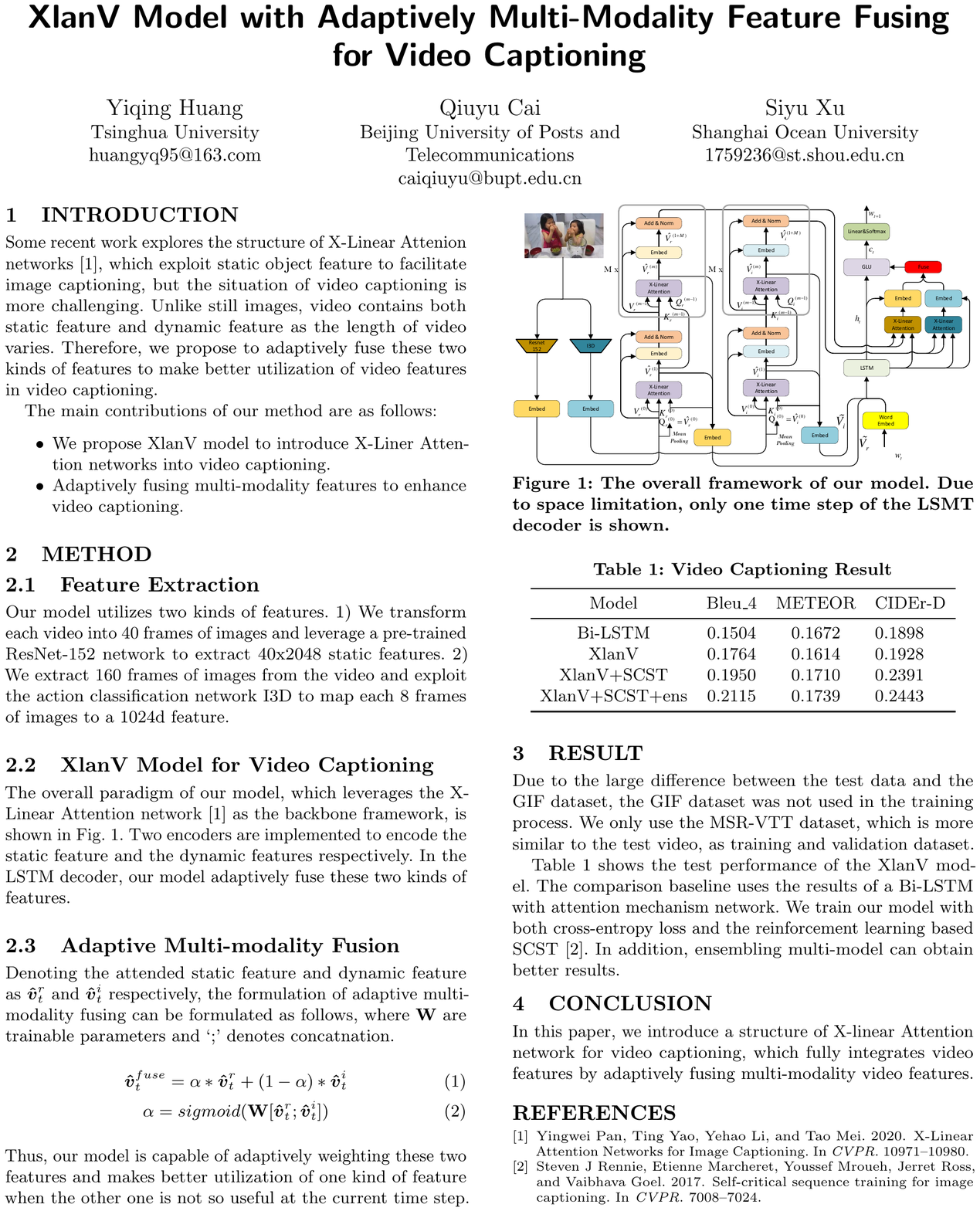}
\end{center}
\newpage
\begin{center}
\includegraphics[width=1\textwidth]{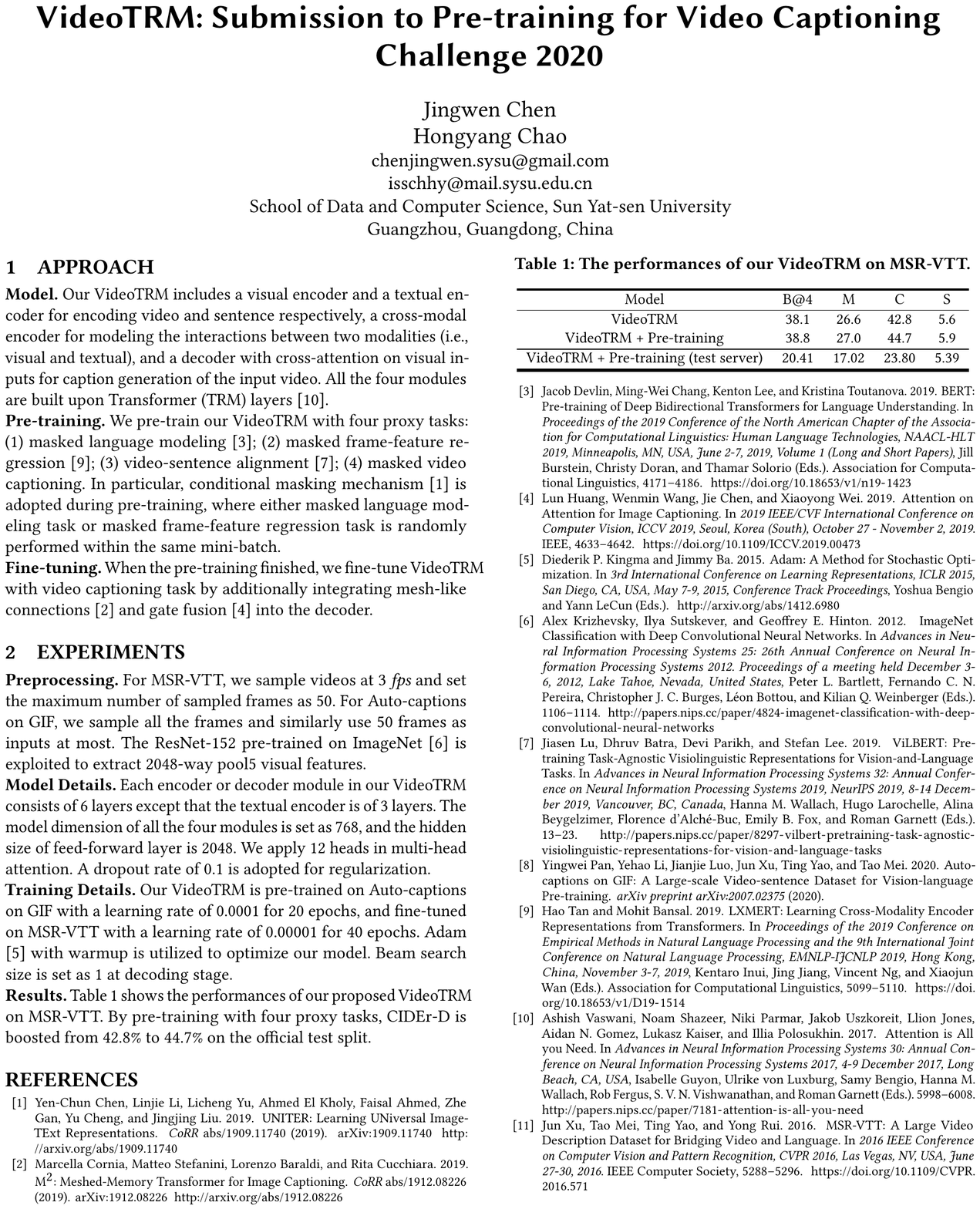}
\end{center}
\newpage
\begin{center}
\includegraphics[width=1\textwidth]{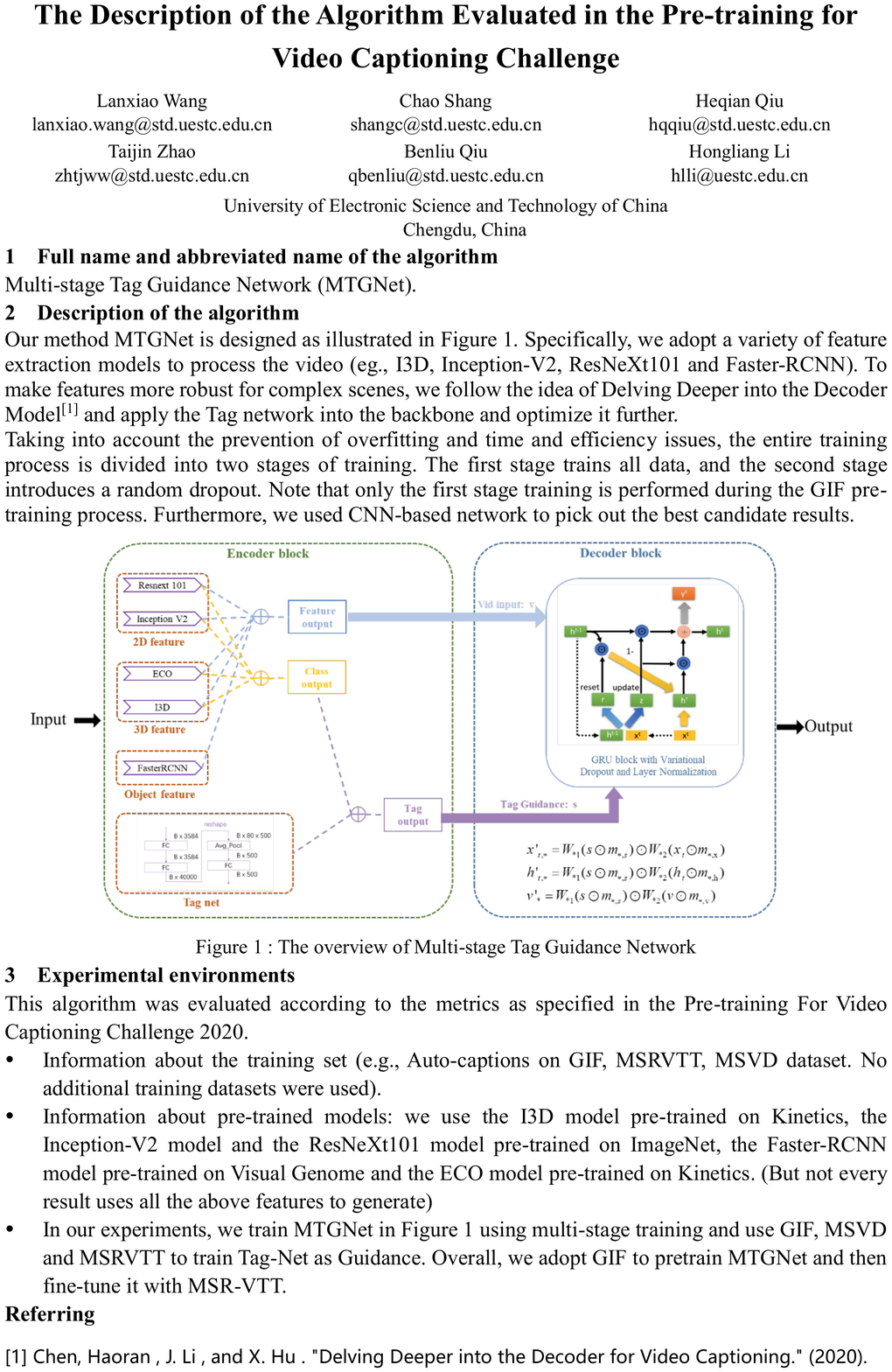}
\end{center}

\end{document}